# Movie Gen: SWOT Analysis of Meta's Generative AI Foundation Model for Transforming Media Generation, Advertising, and Entertainment Industries

Abul Ehtesham
*The Davey Tree Expert Company*
abul.ehtesham@davey.com

Saket Kumar
*The Mathworks Inc*
saketk@mathworks.com

Aditi Singh
*Department of Computer Science*
*Cleveland State University*
a.singh22@csuohio.edu

Tala Talaei Khoei
*Khoury College of Computer Science*
*Roux Institute at Northeastern University*
t.talaeikhoei@northeastern.edu

*Abstract*—Generative AI is reshaping the media landscape, enabling unprecedented capabilities in video creation, personalization, and scalability. This paper presents a comprehensive SWOT analysis of Meta's Movie Gen, a cutting-edge generative AI foundation model designed to produce 1080p HD videos with synchronized audio from simple text prompts. We explore its strengths, including high-resolution video generation, precise editing, and seamless audio integration, which make it a transformative tool across industries such as filmmaking, advertising, and education. However, the analysis also addresses limitations, such as constraints on video length and potential biases in generated content, which pose challenges for broader adoption. In addition, we examine the evolving regulatory and ethical considerations surrounding generative AI, focusing on issues like content authenticity, cultural representation, and responsible use. Through comparative insights with leading models like DALL-E and Google Imagen, this paper highlights Movie Gen's unique features, such as video personalization and multimodal synthesis, while identifying opportunities for innovation and areas requiring further research. Our findings provide actionable insights for stakeholders, emphasizing both the opportunities and challenges of deploying generative AI in media production. This work aims to guide future advancements in generative AI, ensuring scalability, quality, and ethical integrity in this rapidly evolving field.

## I. Introduction

Media generation has always been an art that demands creativity, technical expertise, and significant resources. Traditionally, creating high-quality media content such as videos, music, or cinematic effects require teams of professionals, specialized equipment, and extensive post-production efforts. Recent advances in artificial intelligence, particularly in the field of generative AI [1], have introduced a paradigm shift, making it possible to automate and customize media production in ways that were unimaginable just a few years ago.

Generative AI, which learns patterns from large language model (LLM) datasets [2][3], has evolved rapidly from initially focusing on text generation to encompassing multimodal outputs, including images, audio, and video [3]. This evolution has been marked by impressive advancements in video generation and multimodal integration, enabling AI models to generate synchronized media content across different formats seamlessly. Generative AI models such as DALL-E [4], Sora [5], Google Imagen [6], Phenaki [7], NUWA [8], and VideoStudio [9] are leading the way in generating consistent multi-scene videos, while models like VideoINSTA [10] excel in zero-shot long video understanding through informative spatial-temporal reasoning [11] with LLMs.

Additionally, Video-XL [12] and Loong [13] push the boundaries of generating longer, minute-level videos using autoregressive language models. Meta's Movie Gen [14] stands out by offering highly integrated capabilities for video and audio generation. These innovations are revolutionizing industries such as entertainment, education, and advertising by enabling highly personalized and immersive content, streamlining production processes, and expanding the creative possibilities for content creators. This shift offers new opportunities for customization, efficiency, and the democratization of high-quality media production.

Meta's Movie Gen, a set of foundation models designed for media generation, represents a significant milestone in this evolution. Movie Gen aims to redefine the media production landscape by generating high-quality, 1080p HD videos and synchronized audio from simple text prompts. Its functionality extends beyond traditional media generation; it offers capabilities for personalized video creation, precise video editing, and seamlessly integrated cinematic sound effects. The core innovation lies in its ability to translate complex text prompts into coherent, visually appealing videos, making it a powerful tool for industries such as filmmaking, advertising, education, and immersive experiences.

The architecture of Movie Gen builds upon transformer-based models, inspired by LLaMa3 [15], and incorporates several key modifications for enhanced media generation. With a 30-billion parameter Movie Gen Video model and a 13-billion parameter Movie Gen Audio model, Movie Gen leverages state-of-the-art [2] techniques such as temporal autoencoders [16], progressive resolution scaling, and sophisticated parallelism for training and inference efficiency. The result is a system capable of producing short, yet high-quality video clips complete with rich audio, offering a creative tool for content creators ranging from individual influencers to large production studios.

Despite its cutting-edge capabilities, Movie Gen also presents certain limitations. Currently, it can generate videos of up to 16 seconds in length, which limits its applicability for full-length content production. Moreover, like other generative models, it is prone to biases inherited from its training data, raising concerns about cultural representations and ethical implications. Additionally, issues such as temporal consistency in longer videos and synchronization challenges with intricate audio details indicate areas that require further development. Nevertheless, the post-training procedures [14] incorporated into Movie Gen have led to remarkable capabilities for personalization and precise editing, positioning it as an indispensable tool for creators looking for efficiency and customization.

The SWOT (Strengths, Weaknesses, Opportunities, Threats) framework is a valuable tool for assessing the viability of new technologies, particularly those as transformative as Movie Gen. By evaluating its strengths, such as the quality of generated content and versatility, alongside its weaknesses, opportunities, and threats, we can better understand the broader implications of generative AI for the media industry. This analysis helps identify areas of growth, address challenges, and determine how Movie Gen fits into the competitive generative AI landscape.

The media industry is witnessing rapid changes driven by evolving technologies like generative AI. Movie Gen represents a significant technological breakthrough, enabling new forms of creative expression and reducing the barriers traditionally associated with high-quality content production. Its versatility in text-to-video synthesis, video personalization, video editing, and video-to-audio generation establishes it as a state-of-the-art solution in the domain of media generation. At the same time, these advancements bring both opportunities and challenges—particularly in terms of ethical use, regulation, and competition in the market. In this paper presents a SWOT analysis of Movie Gen, detailing its strengths, weaknesses, opportunities, and threats, providing a comprehensive view of its role and impact in the evolving world of AI-driven media production. Through this analysis, we aim to understand the potential benefits and drawbacks of deploying generative AI in media creation, exploring its capacity to revolutionize creative industries while addressing the technical and ethical hurdles it must overcome.

The contribution of this paper are as follows:

- An in-depth evaluation of Movie Gen's strengths, weaknesses, opportunities, and threats, analyzing its technological and competitive positioning.
- Exploration of Movie Gen's potential in entertainment, advertising, and education, highlighting its adaptability for diverse use cases.
- Comparative Insights on Generative AI Models like DALL-E, Sora, and Google Imagen to identify its unique features and strengths.
- Identification of ethical and technical limitations, including biases, temporal consistency, and synchronization issues.

## II. MOVIE GEN - OVERVIEW

Meta's Movie Gen is a set of foundation models that generate high-quality, 1080p HD videos with synchronized audio. Its capabilities extend beyond simple video generation to include personalized content and precise, instruction-based editing. These models achieve state-of-the-art performance in text-to-video synthesis, personalization, video editing, and audio generation as shown in Fig 1.

### A. Foundation Models

- Movie Gen Video: A 30-billion parameter transformer-based model used for generating video of up to 16 seconds, supporting tasks such as text-to-video synthesis and video personalization
- Movie Gen Audio: A 13-billion parameter model designed to generate synchronized cinematic audio, enhancing the quality of generated video content.

### B. Architecture Components

Transformer Backbone: Based on transformer model [17] inspired by LLaMa3, Movie Gen incorporates advanced modifications like full bi-directional attention, RMSNorm [18], and SwiGLU [19] activation functions for optimal media generation.

Temporal Autoencoder (TAE) compresses RGB images and videos into a spatio-temporally compressed latent representation, significantly reducing the sequence length. This enables efficient processing of both visual and audio data, allowing for high-resolution video generation at native frame rates.

Joint Training for Image and Video Generation: The architecture is designed to handle both text-to-image and text-to-video generation tasks. By treating images as single-frame videos, the model can seamlessly process both data types. The training is conducted in stages, starting with pre-training on images and then jointly training on both images and videos.

### C. Training Techniques

Progressive Resolution Scaling : Training is performed progressively, starting from lower resolutions (256px) to higher resolutions (768px), enhancing both efficiency and generalization.

Post-Training for Additional Capabilities: Personalization and video editing capabilities are added through a post-training procedure. This involves training the model to handle both text and image inputs for personalization and using novel unsupervised methods for precise video editing.

### D. Inference Optimizations

The model employs various inference optimizations to improve efficiency, such as the linear-quadratic time-schedule for faster output generation. Instead of using a traditional linear schedule, the model first takes several linear steps followed by larger, quadratically spaced steps. This maintains output quality while significantly reducing the number of inference steps required.

### E. Scaling and Parallelism

The training process utilizes large-scale infrastructure, including 6,144 H100 GPUs, and incorporates 3D parallelism[14], tensor parallelism[20], sequence parallelism[21], context parallelism[22], and fully sharded data parallelism[23] to efficiently scale the model across multiple GPUs.

Movie Gen also supports video personalization, where it generates personalized videos featuring a specific individual based on an image. Additionally, the model offers precise video editing, allowing users to make specific edits to both real and generated videos using text-based instructions. The emphasis on simplicity in design, drawing from transformer architectures like LLaMa3, combined with the use of temporal autoencoders and efficient training recipes, allows Movie Gen to achieve high-quality media generation across different modalities.

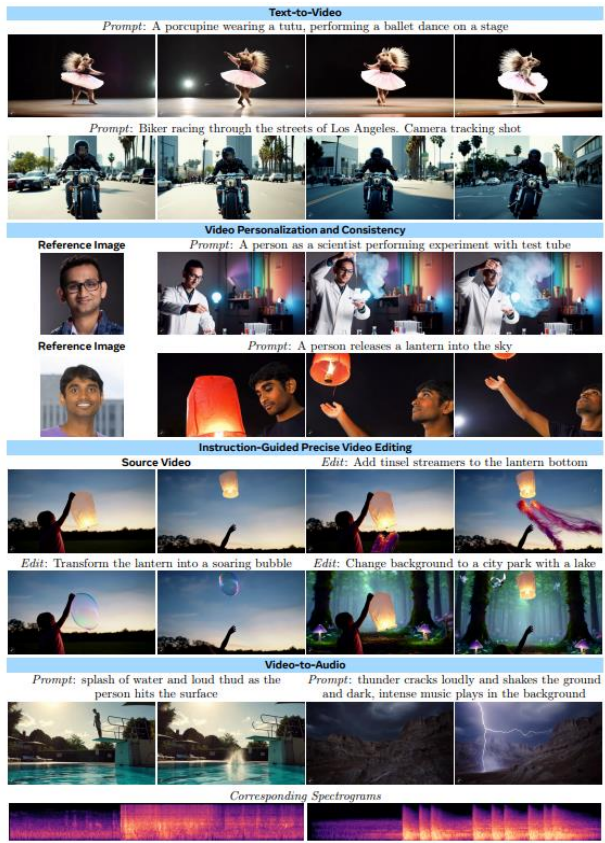

Figure 1. Overview of Movie Gen Capabilities[14]

III. STRENGTH OF MOVIE GEN

*A. High-Quality 1080p HD Video Generation*

Movie Gen sets a new benchmark for high-quality video generation by producing 1080p HD videos with synchronized audio across multiple aspect ratios. The use of a 30-billion parameter transformer and a robust temporal autoencoder (TAE) enables efficient high-resolution generation of up to 16-second videos at 16 frames per second (FPS). Additionally, the Spatial Upsampler [14] is used to increase spatial resolution to full HD while maintaining image quality, making the model ideal for creative industries such as filmmaking and advertising.

*B. Versatile Functionality Beyond Video Generation*

Unlike many competing models, Movie Gen provides comprehensive tools for content creators. It supports:

- Precise Video Editing: Movie Gen offers fine-grained, instruction-guided video editing capabilities that allow users to apply creative modifications to videos using textual prompts. This is achieved without the need for large-scale supervised data by leveraging unsupervised training techniques.
- Video Personalization: Through a post-training procedure [14], Movie Gen allows the generation of personalized videos featuring a person based on their reference image. This enables users to create customized content with a high level of detail and identity preservation.
- Video-to-Audio and Text-to-Audio Synthesis: The Movie Gen Audio model [14], with 13 billion parameters, can generate cinematic sound effects and music synchronized with video input. This is achieved using pre-trained text encoders, which support variable-length audio generation that matches the mood and actions depicted in the visuals.

*C. Scalable Architecture and Efficient Training*

Movie Gen uses an efficient transformer with progressive resolution scaling, employing 3D parallelism for training efficiency. This setup supports large-scale data handling while maintaining generalization across different media formats.

*D. Advanced Training Data and Curation*

The model's success is largely attributed to its extensive and well-curated training dataset, which includes billions of images, millions of videos, and curated audio data. The data curation pipeline includes stages like visual filtering, motion filtering, content filtering, and detailed captioning to ensure high-quality and diverse training material. These steps enhance Movie Gen ability to generate realistic and consistent content while broadening its domain-specific knowledge. The joint training of image and video datasets also improves the model's ability to handle diverse visual concepts.

*E. Instruction-Guided Editing and Personalization*

The post-training procedures for personalization and editing give Movie Gen unique capabilities to generate videos tailored to individual preferences and to perform precise editing tasks with ease. For personalization, Movie Gen can use both textual instructions and reference images to generate a customized video while preserving the unique identity of individuals. This makes Movie Gen a powerful tool for industries looking to create bespoke content without relying on extensive manual editing, thereby significantly reducing production time and cost.

*F. Inference Efficiency and Optimizations*

Movie Gen incorporates an efficient inference pipeline with the linear-quadratic time-schedule for faster video generation. This unique sampling method combines the first few linear steps with subsequent quadratic steps, allowing the model to achieve high-quality output while significantly reducing the number of inference steps. Additionally, multi-diffusion techniques enhance temporal consistency across video frames, minimizing boundary artifacts when generating longer videos, which is crucial for maintaining realism in storytelling and visual narratives.

*G. Reliable Evaluation Framework*

Movie Gen includes a detailed evaluation framework, the Movie Gen Video Bench, which is significantly larger and more thorough compared to existing benchmarks. This framework includes over 1,000 prompts that evaluate text alignment, visual quality, realism, and aesthetics, ensuring the generated videos are of the highest quality across multiple axes. Human evaluators assess these generated outputs in detail, providing more nuanced feedback to guide model improvements.

*H. Model Averaging for Enhanced Performance*

Movie Gen uses a model averaging technique to combine the strengths of multiple finetuned models, which results in improved motion quality, consistency, and camera control. This approach ensures that the final model version is robust across various finetuning data versions, enhancing overall performance and stability during training.

## IV. WEAKNESS OF MOVIE GEN

*A. Video Length Limitations*

The current version of Movie Gen generates short clips suitable for certain advertising needs, but it significantly limits its applicability for full-length content production. As a result, it is unsuitable for generating feature films or other long-duration media.

*B. Bias in Generated Content*

Like other generative AI systems, Movie Gen outputs can reflect biases present in its training data. Despite efforts in data curation, these biases can still lead to culturally skewed or unrealistic portrayals, which poses ethical challenges for content creators, particularly when attempting to represent diverse populations or maintain objectivity.

*C. Evaluation Dependency on Human Judgment*

The evaluation process for generated videos heavily relies on subjective human judgment, which can be inconsistent and influenced by individual biases. Although Meta has attempted to standardize the evaluation with detailed guidelines, the inherent subjectivity and variability in human evaluation mean that assessing content quality remains challenging. Furthermore, existing automated metrics, such as Fréchet Video Distance (FVD) [25] have not proven to reliably correlate with video quality, further complicating consistent evaluations.

*D. Audio Synchronization Challenges*

While Movie Gen excels at generating synchronized audio for basic and straightforward motions, it struggles with more intricate synchronization tasks. This is especially true for complex, visually obscured, or unconventional movements, where maintaining perfect synchronization between video and audio becomes challenging. Moreover, the absence of voice generation capabilities restricts Movie Gen use in applications involving dialogue, limiting its versatility for storytelling or conversational media.

*E. Limited Motion and Realism Consistency*

The model sometimes struggles to produce natural and realistic motion, especially for out-of-distribution concepts or unusual subjects like mythical creatures or surreal settings. The generated motion can appear unrealistic or uncanny, particularly in cases involving detailed limb movements or complex physical interactions. This limitation hinders its effectiveness for scenarios demanding a high degree of realism.Temporal Consistency in Long Videos

Due to memory constraints, Movie Gen employs techniques like temporal tiling for processing long videos during inference. While this helps with scalability, it can lead to inconsistencies at the boundaries of these tiles, resulting in visible artifacts or temporal discontinuities between frames. Although techniques like Multi-Diffusion [24] are employed to reduce these artifacts, such solutions are not always completely effective, especially for long-form content.

## V. OPPORTUNITIES FOR MOVIE GEN

*A. Revolutionizing Filmmaking*

Movie Gen presents significant opportunities to transform the filmmaking process by automating key production stages such as storyboarding, pre-visualization, and special effects generation. The model's capabilities for precise video editing using text-based instructions can help filmmakers rapidly iterate visual effects and scene modifications without requiring extensive manual labor. This ability to generate consistent high-quality visuals and personalized elements allows filmmakers to significantly reduce production costs and time, enabling them to focus on more creative aspects of storytelling.

*B. Targeted Advertising and Marketing*

Movie Gen advanced text-to-video generation capabilities offer substantial potential in advertising and marketing by allowing advertisers to create highly targeted, personalized content that resonates with specific audience segments. The ability to generate HD videos with synchronized audio, alongside personalization features that adapt to a user's profile, positions Movie Gen as a powerful tool for marketers to boost customer engagement and conversion rates. This could lead to an enhanced ability to quickly produce localized or individualized video ads that cater directly to audience demographics and interests, thus revolutionizing digital advertising strategies.

*C. Democratizing Content Creation*

The user-friendly interface and capabilities of Movie Gen democratize the production of professional-quality content. Individual creators, small studios, and influencers can leverage Movie Gen to create videos that would have traditionally required large production teams and specialized skills. This accessibility fosters greater innovation and diversity in content creation, allowing new and diverse voices to participate in the creative process. The model's scalability and efficiency also mean that high-quality outputs are available even to creators who lack extensive computational resources, ultimately breaking down barriers to entry in video production.

*D. Educational and Personalized Learning Content*

The personalization capabilities of Movie Gen provide a significant opportunity in educational technology. By enabling the creation of customized videos tailored to individual students, Movie Gen can help educators deliver personalized learning experiences that cater to unique learning styles and preferences. Teachers can generate interactive and engaging educational content that includes the student's name or likeness, increasing relatability and engagement. Additionally, Movie Gen ability to synchronize video with audio enables immersive educational videos that enhance the learning experience, which is particularly valuable for younger audiences or those with different learning needs.

*E. Audio Generation and Sound Design Enhancements*

Movie Gen potential for advancements in audio generation includes opportunities to expand into voice synthesis, better synchronization for more intricate video scenes, and enhanced sound effects. The current capabilities of Movie Gen Audio for synchronized sound effects and music could be extended to include voiceovers, allowing for full dialogue-driven content. This expansion would provide content creators with an all-in-one tool for cinematic sound design, further streamlining the production pipeline for movies, advertisements, and digital media. Voice synthesis capabilities, combined with improved video personalization, could also make Movie Gen useful for creating personalized storytelling content for children or promotional materials that require direct viewer engagement.

### F. Immersive Media and Virtual Experiences

Movie Gen has the potential to revolutionize virtual experiences and immersive media. By generating realistic visuals and synchronized sound, Movie Gen can be used to create content for virtual reality (VR) and augmented reality (AR) applications, providing users with rich and engaging experiences. This could be used in various sectors, including gaming, virtual tourism, and remote events, creating new possibilities for experiential media. The high-quality video and audio generation can help enhance the realism and immersion of these experiences, making them more compelling and interactive.

### G. Collaborations with Creative Industries

The entertainment, fashion, and music industries present rich opportunities for collaboration with Movie Gen AI. Designers and musicians could use the model to rapidly generate promotional visuals, music videos, or animated runway shows. The ability to edit videos based on textual descriptions allows creatives to experiment with visual effects and settings without needing technical expertise, fostering a more iterative and imaginative creative process. Moreover, Movie Gen video-to-audio generation capability allows artists to explore unique audiovisual effects, adding depth to visual art and creating immersive, synesthetic experiences.

### H. Potential for Real-Time Video Generation

With further advancements in inference optimizations and GPU infrastructure, Movie Gen could potentially evolve to support real-time video generation. This would be a game-changer for live events, allowing personalized media to be created instantly in response to audience interactions. This opportunity is particularly relevant for entertainment and educational live events, where interactive, dynamic content could be generated on the fly to enhance audience engagement.

## VI. THREATS FACING MOVIE GEN

### A. Ethical Concerns and Potential for Misuse

Movie Gen ability to generate realistic, high-quality videos presents serious ethical challenges, particularly around the misuse of its capabilities for deep-fake technology and misinformation. The advanced personalization features and video editing capabilities, while highly innovative, can be exploited to create misleading or harmful content. The ability to easily generate and edit videos with convincing quality means that misuse could lead to significant reputational harm, privacy violations, or political misinformation campaigns. This necessitates the implementation of robust safeguards, strict usage policies, and ethical guidelines to mitigate the potential misuse of the technology for malicious purposes.

- Bias Amplification: Movie Gen models can inherit biases present in the training data. For example, if the training data contains predominantly videos featuring certain demographics or portraying specific cultural perspectives, the model may generate content reflecting those biases, potentially exacerbating societal prejudices.
- Unintentional Associations: The models can learn unintentional associations between text prompts and generated media, leading to outputs that may be misconstrued or misused. This underscores the importance of robust content filtering and monitoring mechanisms to mitigate the risk of generating inappropriate or harmful content.
- Lack of Transparency in Commercial Systems: Many commercial text-to-video and text-to-audio systems operate as black boxes, which hinders the ability to fully understand their limitations, potential biases, and the data used for training. This lack of transparency makes it challenging to address ethical concerns comprehensively.

### B. Public Perception of AI-Generated Content

The acceptance of AI-generated content will largely depend on public perception, which poses a significant threat to the wide adoption of Movie Gen. The model's capabilities to create highly realistic videos, including personalized and edited content, might raise concerns around authenticity, copyright, and the potential to replace human creativity. Additionally, the biases inherent in the training data, despite extensive filtering and data curation efforts, could negatively affect public trust. The generated content could unintentionally reinforce stereotypes or culturally insensitive portrayals, which may harm the adoption of the technology, especially within creative industries that value authenticity and originality.

- Authenticity and Trust: Public acceptance of AI-generated content hinges on trust and the perceived authenticity of the media. Concerns about deepfakes, misinformation, and the potential for malicious use can erode trust in AI-generated content.
- Impact on Human Creativity: There is a debate regarding the potential impact of AI on human creativity. Some argue that AI tools can augment and empower creators, while others fear that AI-generated content may devalue human artistic expression and lead to job displacement in creative industries. Addressing these concerns and demonstrating the value of AI as a tool for creative collaboration will be essential.

### C. Reliance on Human Evaluation

Movie Gen AI current evaluation process is heavily reliant on human judgment to assess the quality of generated videos. Although detailed guidelines are provided to evaluators, human biases, subjective preferences, and inconsistencies pose a threat to the reliability of assessments. Automated metrics like FVD (Fréchet Video Distance) [25] have proven insufficient in correlating with human evaluations for assessing video quality. Therefore, the dependency on human evaluators limits the scalability of the evaluation process and the ability to ensure consistent quality in the long term. The development of reliable automated metrics is needed to reduce these threats and scale the evaluation process effectively.

### D. Legal and Regulatory Challenges

The regulatory environment for AI-generated content is evolving, and Movie Gen could face legal challenges regarding intellectual property, content ownership, and liability for misuse. As different jurisdictions enact regulations on deepfake technology and generative AI, Movie Gen must navigate a complex and changing legal landscape to ensure compliance. This poses a potential risk for the technology's adoption and limits its deployment across different markets. Failure to address these regulatory challenges proactively could result in restrictions, fines, or limitations on the commercial viability of Movie Gen.

VII. COMPARATIVE ANALYSIS WITH COMPETING GENERATIVE AI TOOLS

Movie Gen operates in a highly competitive landscape, with several generative AI tools aiming to transform media content creation. This section compares Movie Gen with its key competitors, including OpenAI's Sora [5], Runway Gen3 [26], LumaLabs[27], AmazonNova[29]-as detailed in Table 1.

- **OpenAI's Sora:** A high-quality generative AI model for video and audio generation, focusing on creative flexibility and narrative style. Sora excels in user-controlled content but lacks the tight audio-video integration of Movie Gen.
- **Runway Gen3**: A text-to-video model catering to artists and professionals with advanced video editing tools and support for longer durations. While versatile, it lacks the personalization capabilities of Movie Gen, which makes the latter better for advertising and tailored entertainment.
- **LumaLabs**: Specialized in 3D video and AR/VR content generation, it offers spatially dynamic experiences. While exceptional for immersive media, LumaLabs falls short in traditional 2D video editing and audio integration compared to Movie Gen.
- **Amazon Nova Reel**: A diffusion-based model focusing on high-quality 720p text-to-video generation with consistent motion and visuals. It offers a strong foundation but lacks integrated audio generation and advanced personalization features like Movie Gen.

Table 1. Comparative Analysis

| Feature/Aspect | OpenAI SORA [5] | RunwayML Model[26] | Luma Model[27] | Amazon Nova Reel[29] | Meta Movie Gen[14] |
|---|---|---|---|---|---|
| **Primary Focus** | Text-to-video | Image-to-video / Text-to-video | Image-to-video | Text-to-video with optional image input | Text-to-video, video personalization, and video editing |
| **Architecture Type** | Transformer-based | GANs (Generative Adversarial Networks) [28] | Neural Radiance Fields (NeRF) [29] | Diffusion-based | Transformer-based with Temporal Autoencoder (TAE) |
| **Key Components** | Transformer decoder Cross-attention | GAN-based encoder-decoder pipeline | NeRF-based voxel grid architecture | Diffusion process conditioned on text and image input | Transformer with Flow Matching objective TAE for spatio-temporal compression |
| **Training Dataset** | Large-scale text-image/video dataset | Custom image-video pairs | 3D scenes with multi-view data | Curated prompts for human activities, natural scenes | 100M video-text pairs; 1B image-text pairs |
| **Model Parameters** | Unknown | Unknown | Unknown | Unknown | 30B parameters (Movie Gen Video model) |
| **Output Quality** | High-quality text-to-video with temporal coherence | High-quality video with style control | Realistic, smooth 3D video | High-quality 720p videos, superior consistency | 1080p HD videos with synchronized audio Supports aspect ratio adjustments |
| **Temporal Consistency** | Excellent | Moderate | High (due to 3D focus) | High temporal coherence for subjects/backgrounds | Excellent (using spatio-temporal latent compression) |
| **Rendering Approach** | Frame-based generation | Frame-by-frame with style adaptation | Continuous 3D rendering | Frame-based generation with image/motion focus | Spatio-temporal compressed latent generation |
| **Video Personalization** | Limited | Limited | Limited | Not specified | Strong support for personalized video generation based on a reference image |
| **Video Editing** | Limited | Limited | Limited | Not specified | Instruction-based precise video editing |
| **Sound and Audio Generation** | Not integrated | Not integrated | Not integrated | Not integrated | Integrated video-to-audio and text-to-audio generation |
| **Training Approach** | Pre-trained on text-image/video pairs | Adversarial training | NeRF-trained on multi-view images | Diffusion-based model | Pre-training on large video-text pairs Finetuning on curated videos |
| **Strengths** | High-quality coherent video generation | Artistic flexibility | High realism in 3D environments | Superior video quality and consistency | High-resolution video (up to 1080p), video editing, and personalization |
| **Weaknesses** | Computationally expensive | Temporal consistency issues in complex videos | Focused on 3D only | Limited video length/resolution details | Requires extensive training resources (6,144 GPUs used for training) |

## VIII. CONCLUSION

Meta's Movie Gen stands at the forefront of media generation technologies, demonstrating significant potential to transform the filmmaking, advertising, and educational sectors through automated content creation. Our SWOT analysis highlights Movie Gen robust capabilities in HD video generation, personalization, and audio synchronization, while acknowledging limitations such as video length constraints, bias in generated content, and challenges related to temporal consistency.

Despite these challenges, Movie Gen presents unique opportunities for democratizing content creation, enhancing personalized learning, and enabling immersive media experiences. However, technology also faces threats from ethical misuse, regulatory changes, and the fast-paced evolution of generative AI competitors. To maintain its competitive edge and positive societal impact, continued research, ethical safeguards, and innovative development are essential for Movie Gen AI's growth. As generative AI reshapes the landscape of content production, Movie Gen is well-positioned to lead this transformation, provided it navigates the associated risks responsibly.